\documentclass[preprint,12pt]{elsarticle}
\usepackage{graphicx}
\graphicspath{{./figures/}}
\usepackage{natbib}
\usepackage{picinpar}
\usepackage{amsmath, amsfonts}
\usepackage[caption=false,font=footnotesize]{subfig}
\usepackage{url}
\usepackage{flushend}
\usepackage{colortbl}
\usepackage{soul}
\usepackage{multirow}
\usepackage{pifont}
\usepackage{color}
\usepackage{alltt}
\usepackage[hidelinks]{hyperref}
\usepackage{enumerate}
\usepackage{siunitx}
\usepackage{breakurl}
\usepackage{epstopdf}
\usepackage{pbox}
\usepackage{makecell}

\begin{document}

\begin{frontmatter}

\title{A Deep Learning Framework for Simulation and Defect Prediction Applied in Microelectronics}

\author[label1]{Nikolaos Dimitriou\corref{cor1}}
\ead{nikdim@iti.gr}
\cortext[cor1]{Corresponding author.}

\author[label1]{Lampros Leontaris}

\author[label1]{Thanasis Vafeiadis}

\author[label1]{Dimosthenis Ioannidis}

\author[label2]{Tracy Wotherspoon}

\author[label2]{Gregory Tinker}

\author[label1]{Dimitrios Tzovaras}

\address[label1]{Information  Technologies  Institute,  Centre  for Research  and Technology  Hellas}
\address[label2]{Microsemi Corporation}
	
\begin{abstract}
The prediction of upcoming events in industrial processes has been a long-standing research goal since it enables optimization of manufacturing parameters, planning of equipment maintenance and more importantly prediction and eventually prevention of defects. While existing approaches have accomplished substantial progress, they are mostly limited to processing of one dimensional signals or require parameter tuning to model environmental parameters. In this paper, we propose an alternative approach based on deep neural networks that simulates changes in the 3D structure of a monitored object in a batch based on previous 3D measurements. In particular, we propose an architecture based on 3D Convolutional Neural Networks (3DCNN) in order to model the geometric variations in manufacturing parameters and predict upcoming events related to sub-optimal performance. We validate our framework on a microelectronics use-case using the recently published ``PCB scans'' dataset where we simulate changes on the shape and volume of glue deposited on an Liquid Crystal Polymer (LCP) substrate before the attachment of integrated circuits (IC). Experimental evaluation examines the impact of different choices in the cost function during training and shows that the proposed method can be efficiently used for defect prediction.
\end{abstract}

\begin{keyword}
deep learning, defect prediction, industrial simulation, integrated circuit.
\end{keyword}

\end{frontmatter}

\section{Introduction}

Prediction of upcoming failures is critical for industrial processes as it enables prevention of defects as well as timely intervention, namely equipment maintenance and calibration. Therefore, several methods and approaches have been proposed that can forecast defective or sub-optimal states either from previous measurements or through user-defined models. In this category fall traditional techniques, such as Finite Element Analysis \cite{8641633},\cite{7801118},\cite{6862014}, particle-based methods \cite{7060661} and time-series \cite{7310861} that have certainly improved the respective processes in terms of robustness and productivity. Simultaneously, the transition to Industry 4.0 and the installation of Internet of Things (IoT) sensory networks on industrial shop-floors \cite{8401919}, \cite{8410462} has enabled the continuous monitoring of production through analysis of productions data.

This adundance of sensory data has also facilitated the proliferation of Deep Learning (DL) in industry which has further advanced predictive analysis with models such as Recurrent Neural Networks (RNN) \cite{recurrent} and its popular Long Short-Term Memory (LSTM) \cite{8039509}, \cite{8353154} and Gated Recurrent Unit (GRU) \cite{7997605} variants being able to model temporal sequences and provide meaningful predictions from previous measurements. Aside prediction, DL has significant impact in another aspect of quality monitoring namely in defect detection. In this research direction, several DL architectures have been proposed for the identification of a defect after it has occurred \cite{8676279}, \cite{8335303}, \cite{8664657}, showing notable results that improve overall production quality and save resources by stopping the propagation of defects at later manufacturing phases. 

In this paper we aim to close the loop between defect detection and prediction by  a 3DCNN for defect classification that is then used as an additional supervisory signal in training a 3DCNN that simulates the variations that occur in a critical, in terms of quality, 3D structure. As is experimentally validated this additional supervision is crucial in generating simulations that are qualitatively close to the actual 3D structures and more importantly can be used for the prediction of defects. An interesting aspect of the paper is that contrary to previous work on forecasting that focuses on time series data or 2D images, the proposed framework is developed and evaluated on 3D data. To summarize, the main novelties of our work can be pinpointed to:
\begin{itemize}
\item an adaptation of the RNet \cite{RNet} regression network that is directly used for defect detection.
\item a DL defect prediction architecture that based on previous 3D measurements can extrapolate a quality-critical 3D structure in a future state.
\item a training process that substantially improves prediction results and closes the loop between defect detection and prediction by utilizing a DL defect detection pipeline as supervision during training of the prediction network.
\end{itemize}
The entire framework is experimentally validated on a real scenario in microelectronics industry using the recently published dataset of \cite{PCB_scans} where variations in the glue dispensation process for IC attachment are simulated and defective states are predicted.

The paper structure is as follows; Section \ref{sec:related} is devoted to the presentation of state of the art methods for prediction and forecasting on critical quality parameters in industrial processes. Subsequently in Section \ref{sec:descrip} the proposed DL framework is presented with both models for defect detection and prediction whereas Section \ref{sec::experiments} is devoted to the experimental evaluation of the proposed method. Finally, Section \ref{sec:conclusions} draws the conclusions of our work.

\section{Predictive Algorithms in Industrial Processes}
\label{sec:related}

In the current section we will attempt a brief yet representative literature review of prediction algotithms that have been successfully used in industrial environments. In addition to this, we summarize and group existing methods in Table \ref{tab::rel_categories}. Local feature-based gated recurrent unit (LFGRU) networks have been proposed in \cite{7997605} to predict machine condition by further processing of handcrafted features that aggregate time series information, using a gated recurrent unit network to learn richer feature representations. In order to predict defects and model degradation phenomena in renewable energy storages, the work of \cite{8624619} introduces an error correction factor that enhances the grey model (GM) without increasing complexity. A predictive method for  remaining useful life (RUL) estimation is proposed in \cite{7726039}, that  utilizes support vector regression to directly model the correlation between sensor values or health indicators and estimate RUL of equipment. In a related work \cite{8186223}, on RUL estimation and state diagnosis, a support vector regression model is used to simulate the battery aging mechanism and estimate impedance variables whereas a particle filter is employed to mitigate measurement noise and accurately estimate the impedance degradation parameters. In a related line of research, the problem of RUL estimation and state of health in lithium-ion batteries is investigated in \cite{8310615}, using a degradation model based on traveling distance in Brownian motion and particle filtering for estimating the drift of a Brownian particle. A double-scale particle filtering method is also introduced in \cite{7997600}, to predict battery remaining available energy and estimate state-of-charge (SOC) under temperature uncertainties and inaccurate initial SOC values. In \cite{8445706}, prediction of RUL has been researched for wind turbine drivetrain gearboxes where a particle filtering algorithm is introduced that employs a neuro-fuzzy inference system to model state transition and a multinomial resampling method to tackle particle impoverishment. The authors of \cite{8664594} propose an ensemble classifier using density, geometry, and radon-based features and combining several classification algorithms to identify defect-related wafer map patterns. 

\begin{table*}
\tiny
\centering
\begin{tabular}{|c|c|l|c|}
\hline
\textbf{\textit{Application domain}} & \textbf{\textit{Reference}} & \multicolumn{1}{c|}{\textbf{\textit{Application task}}} & \textbf{\textit{Main methodology}} \\
\hline
\multirow{10}{*}[-5em]{\makecell[cl]{Fault diagnosis \\ detection and \\monitoring}} & Dang et al. \cite{8676279} & \makecell[cl]{Industrial multiphase flow \\ monitoring, oil-water \\ flow} & CNN, LSTM \\
\cline{2-4}
 & Dimitriou et al. \cite{RNet} & \makecell[cl]{Printed Circuit Board defect \\  defection} & 3D CNN \\
\cline{2-4}
 & Saqlain et al. \cite{8664594} & \makecell[cl]{Semiconductor wafer patterns \\ defect detection} & MLC, SVE \\
\cline{2-4}
 & Wen et al. \cite{101} & \makecell[cl]{Bearing fault deterioration \\ diagnosis} & \multirow{7}{*}[-3.5em]{\makecell[cc]{Various forms \\of CNN}} \\
\cline{2-3}
 & Wen et al. \cite{102} & \makecell[cl]{Motor bearing self-priming \\ centrifugal pump and axial \\ piston hydraulic pump \\ fault diagnosis} & \\
\cline{2-3}
 & Sun et al. \cite{sun0} & \makecell[cl]{Induction motor fault \\ diagnosis} & \\
\cline{2-3}
 & Shao et al. \cite{108} & \makecell[cl]{Induction motor, gearboxes \\ and bearing fault \\ diagnosis} & \\
\cline{2-3}
 & Weimer et al. \cite{109} & \makecell[cl]{Defect detection in statistically \\ textured surfaces} & \\
\cline{2-3}
 & Tello et al. \cite{tell0} & \makecell[cl]{Semiconductor manufacturing \\ defect detection} & \\
\cline{2-3}
 & Xie et al. \cite{8664657} & Sewer defect detection & \\
\hline
\multirow{4}{*}[-2.0em]{\makecell[cl]{Long term \\ prediction and \\ forecasting}} & Lipton \cite{recurrent} & Sequence learning & RNN \\
\cline{2-4}
 & Kong et al. \cite{8039509} & \makecell[cl]{Short-term Residential \\ Load Forecasting} & LSTM \\
\cline{2-4}
 & Fang et al. \cite{8353154} & \makecell[cl]{Mobile per-cell \\ demand prediction} & GCN, LSTM \\
\cline{2-4}
 & Hoermann et al. \cite{8460874} & \makecell[cl]{Long term situation \\ prediction, autonomous \\ driving} & 3D CNN \\
\hline
\multirow{10}{*}[-5em]{\makecell[cl]{Remaining \\ useful life \\ and tool wear}} & Zhao et al. \cite{7997605} & \makecell[cl]{Machine health monitoring \\ tool wear prediction} & RNN \\
\cline{2-4}
 & Zhou et al. \cite{8624619} & \makecell[cl]{\makecell[cl]{Lithium-Ion and fuel \\ cell aging prediction}} & \makecell[cc]{Improved Grey \\ Prediction \\  Model} \\
\cline{2-4}
 & Khelif et al. \cite{7726039} & \makecell[cl]{\makecell[cl]{Turbofan engine\\ degradation dataset}} & \multirow{2}{*}{SVR} \\
\cline{2-3}
 & Wei et al. \cite{8186223} & \makecell[cl]{Battery aging prediction} & \\
\cline{2-4}
 & Dong et al. \cite{8310615} & \makecell[cl]{Lithium-Ion batteries \\ degradation prediction} & \multirow{3}{*}[-1.5em]{\makecell[cc]{Particle \\ Filtering}} \\
\cline{2-3}
 & Xiong et al. \cite{7997600} & \makecell[cl]{Remaining available energy \\ and state of charge \\ prediction} &  \\
\cline{2-3}
 & Cheng et al. \cite{8445706} & \makecell[cl]{Drivetrain gearboxes of \\ wind turbines RUL \\ prediction} & \\
\cline{2-4}
 & Khdoudi et al. \cite{104} & \makecell[cl]{Ultrasonic welding \\ parameters prediction} & \multirow{2}{*}[-1em]{CNN}\\
\cline{2-3}
 & Liu et al. \cite{105} & \makecell[cl]{Bearing fault recognition \\ and RUL prediction in \\ parallel} & \\
\cline{2-4}
 & Dimitriou et al. \cite{8448952} & Surface deterioration prediction &  3D CNN\\
\hline
\end{tabular}
\caption{Categorization of predictive algorithms in industry. MLC stands for Machine Learning Classifier,  GCN for Graph Convolutional Network, SVE for Soft Voting Ensemble and SVR for Support Vector Regression.}
\label{tab::rel_categories}
\end{table*}

Fault Diagnosis (FD) is a field with many crucial applications in industrial processes and vital in Industry 4.0. There are many and interesting applications and methodologies proposed by authors in an effort to try and cover the ongoing and growing challenges in this field. In \cite{101}, the authors propose a methodology based on hierarchical convolutional neural networks (HCNN) as a two level hierarchical diagnosis network with two main characteristics: the fault pattern and fault severity are modelled as one hierarchical structure and estimated at the same time. Based on that structure the proposed architecture has two classifiers. In \cite{102}, a convolutional neural network based on LeNet-5 \cite{103} is proposed for fault diagnosis, where the input is the converted signal(s) in a two-dimensional image. In \cite{104}, a machine learning algorithm for the prediction of the suitable machine parameters to achieve good quality for a specific product is proposed based on convolutional neural networks with an application on industrial process parameter prediction. The authors of \cite{105} introduce a joint-loss convolutional neural network approach (JL-CNN) so as to capture common features between FD and RUL problems. The proposed architecture is based on CNN and implements bearing fault recognition and RUL prediction in parallel by sharing the parameters and partial networks, meanwhile keeping the output layers of different tasks. The work in \cite{RNet} proposes a three-dimensional convolutional neural network (3D-CNN) architecture called RNet that automates fault diagnosis by estimating accurately the volume of glue deposits on Printed Circuit Boards (PCB) so as to attach silicon die or other wire bondable components. Additionally, a convolutional discriminative feature learning method is presented for induction motor fault diagnosis in \cite{107}. The approach firstly utilizes back-propagation (BP)-based neural network to learn local filters capturing discriminative information. Then, a feed-forward convolutional pooling architecture is built to extract final features through these local filters. Another interesting approach is presented in \cite{108} where the authors develop a deep learning framework to achieve high performance on machine fault diagnosis using transfer learning so as to enable and accelerate the training of deep learning network. The input of this pipeline is original sensor data that are converted to images by conducting a wavelet transformation to obtain time-frequency distributions. Moreover the authors in \cite{109} examine alternative design configurations of deep convolutional networks and the impact of different parameter settings in the accuracy of defect detection results. 

In more general tasks in industrial processes domain, the authors of \cite{8460874} utilize a dynamic occupancy grid map that is processed by a deep convolutional neural network that models road user's interaction in order to predict complex scenarios in intelligent vehicles navigation. The work of \cite{8448952} uses a 3DCNN to model and predict surface deterioration phenomena on metallic materials used in artwork based on previous 3D scans of artificially aged reference samples. In \cite{110}, the authors propose a methodology based on convolutional neural networks, called RouteNet that performs preventive measures in early routability prediction so as to avoid design rule violation. 

\section{Deep learning framework for simulation and defect prediction}
\label{sec:descrip}

In this section, we describe the proposed approach explaining the data parameterization process as well as the architecture of the deployed deep 3D convolutional networks.

\subsection{Data parameterization}
\label{sec::dataparam}

\begin{figure}
\centering
\subfloat[]{\includegraphics[width=0.3\linewidth]{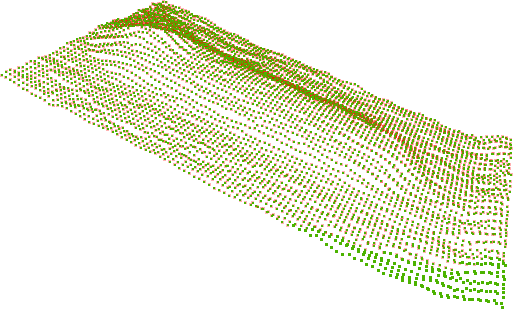}}
\hfil
\subfloat[]{\includegraphics[width=0.3\linewidth]{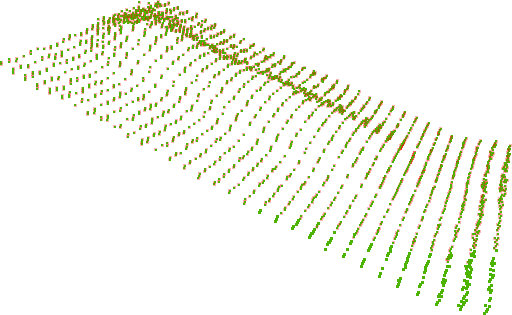}}
\hfil
\subfloat[]{\includegraphics[width=0.3\linewidth]{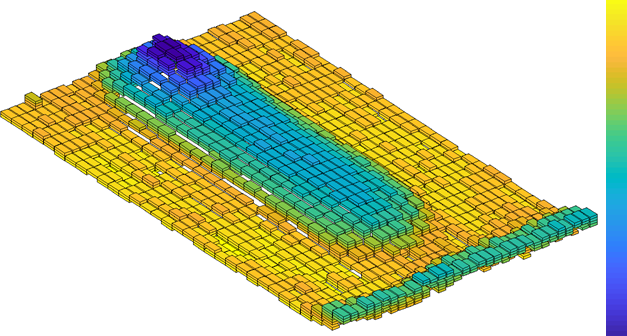}}
\caption{(a) The scanned point cloud of a glue deposit where the captured elevation is the actual mass of glue dispensed over the LCP substrate. The scanned area is $1.2 \times 2.1\ mm^2$. (b) The point cloud converted to an occupancy grid after subsampling to a $32 \times 32 \times 64$ resolution. (c) The same occupancy grid with a parallelepiped centered on each point. The color scale shown on the right shows the elevation on the downward pointing $z$ axis.}
\label{fig:quan}
\end{figure}

In any manufacturing process where we want to monitor a specific quality-sensitive region, the proposed method has as input the 3D point clouds of previous specimens and will provide an estimate of the next one. For instance, in the examined microelectronics use case where we monitor glue deposition on a LCP, the input is four previous point clouds corresponding to previous glue depositions. As explained in \cite{RNet}, \cite{PCB_scans} point clouds in the examined \textit{PCB scans} dataset, are acquired by a scanning system consisting of a laser sensor and two orthogonal linear stages that move the LCP, whereas scanning parameters are constant during data acquisition. Therefore equal areas are covered in each scan while stages step is constant to $20 \mu m$ with micro-meter level accuracy. This allows us to register the 3D point clouds in a common coordinate system. Each point cloud is converted to a $0/1$ occupancy grid having a $32 \times 32 \times 64$ resolution along the $x,y,z$ axes. An example of this process is depicted in Figure \ref{fig:quan} where the scanned point cloud is shown along with the subsampled one after volumetric quantization and the extracted occupancy grid.

\subsection{Network architecture}
\label{sec::nets}

The proposed framework incorporates two 3DCNN models. The classification model characterizes a 3D scan based on the monitored quality parameter. On the other hand, the simulation model produces a point cloud corresponding to the future state of the monitored 3D structure by simulating geometric variations based on previous 3D scans. In the studied use case the input 3D scans are the previous point clouds of the areas where glue is dispensed and the output is the point cloud of the next glue deposition.

\begin{figure}
\centering
\includegraphics[width=0.99\linewidth]{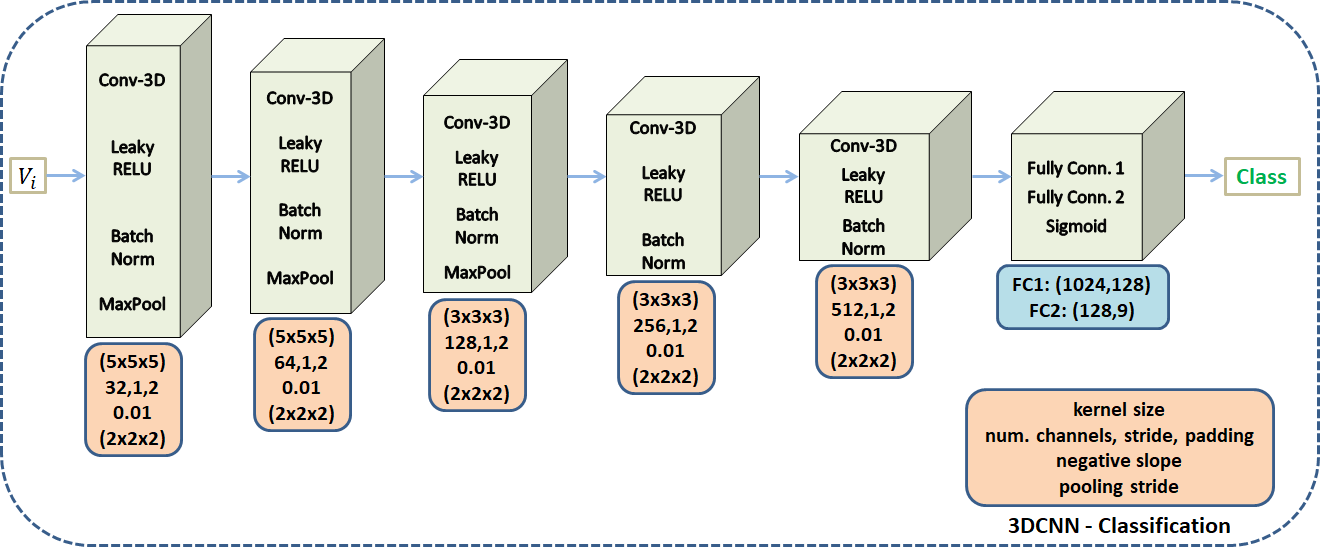}
\caption{The architecture of the classification model. It has five convolutional blocks that reduce the spatial dimensions while increasing the number of channels. Subsequently the network has two fully connected layers followed by sigmoid activation and produces a $9$ elements vector with scores for each class.}
\label{fig:3dcnn-clas}
\end{figure}

The architecture of the classification model along with the parameters of each layer are depicted in Figure \ref{fig:3dcnn-clas}. Essentially the architecture is based on RNet \cite{RNet} where we have removed the final fully connected layer and adjusted the architecture in order to perform classification. Concretely, the network input is an occupancy grid $V_i$ and comprises five convolutional blocks that, identically to \cite{RNet}, progressively decrease spatial dimensions while increasing the number of channels. Each block implements a sequence of layers, namely 3D convolution, leaky ReLU activation, batch normalization and max pooling. After the convolutional blocks the network has two fully connected layers followed by a sigmoid activation that produces a one-hot classification vector $v_i$. Eventually the predicted class corresponds to the maximum entry of $v_i$. For training, Adam optimizer \cite{adam0} is deployed while the cross entropy loss function is used \cite{deeplbook}. More formally the loss function is,
\begin{equation}
L_{C} = -\sum_{i=1}^{B}\sum_{j=1}^{M} y_{ij}log(v_{ij})
\label{eq:cla-loss}
\end{equation} 
where $B$ is the batch size and $M$ denotes the number of classes. Variables $y_i$ and $v_i$ represent respectively the ground-truth and predicted class vector of occupancy grid $V_i$, with $y_{ij}$ having a $\{0,1\}$ value depending on whether $V_i$ belongs to the $j$-th class and $v_{ij}$ being the predicted class score.

\begin{figure*}
\centering
\includegraphics[width=0.99\linewidth]{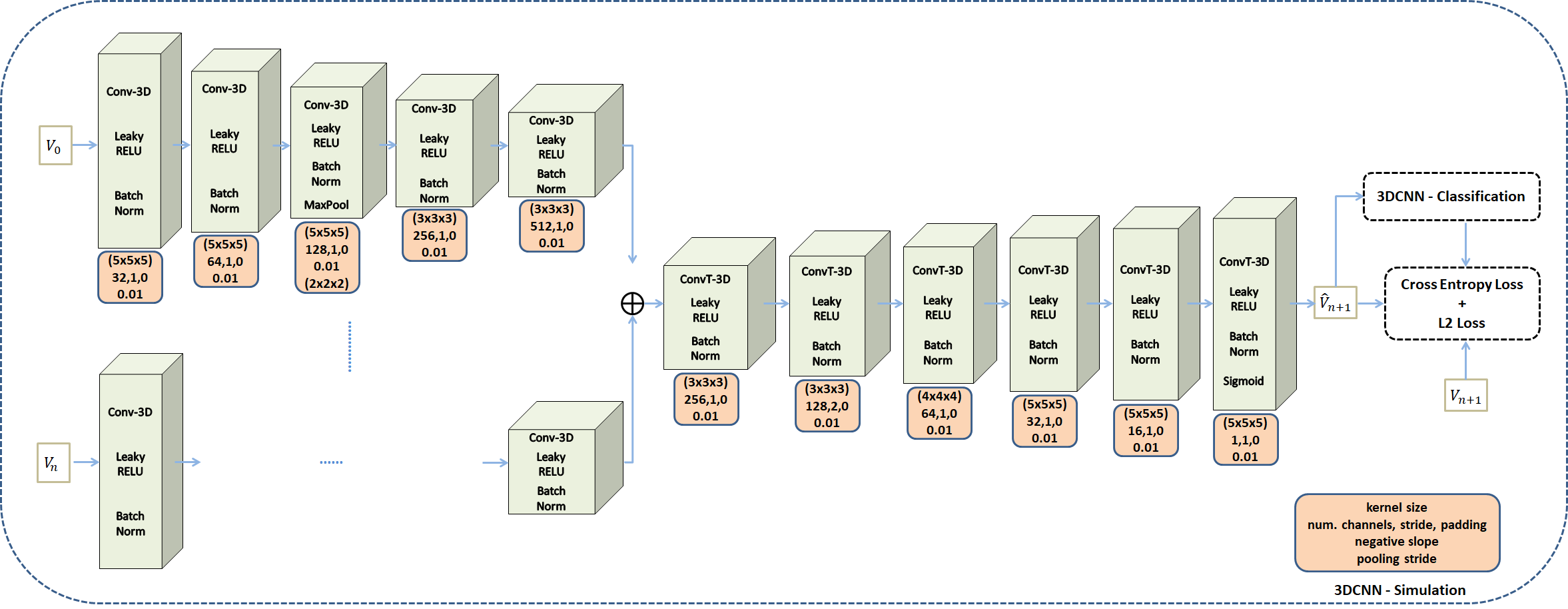}
\caption{The architecture of the simulation model. It has $n+1$ sub-networks consisting of five convolutional blocks each that take as input the past occupancy grids. After spatial decimation, the generated representations are averaged and the result is spatially upsampled to the resolution of an occupancy grid using six transpose convolution blocks. For training, the $L_2$ loss between the ground-truth and predicted occupancy grids is combined with the cross entropy loss between the ground-truth label of the occupancy grid and the predicted one using the classification network of Figure \ref{fig:3dcnn-clas}.}
\label{fig:3dcnn-pred}
\end{figure*}

The simulation model has a more complex architecture as shown in Figure \ref{fig:3dcnn-pred} which also includes the parameters of each layer. Denoting the occupancy grids of previous measurement as $\{V_0,...,V_n\}$, initially each grid $V_i$ is processed separately by diminishing its spatial dimensions while increasing the number of channels. This part of the network consists of five convolutional blocks combining a 3D convolutional layer, leaky ReLU activation and batch normalization while the middle block has a max pooling layer as well. The initial spatial dimensions of $32 \times 32 \times 64$ drops to $6 \times 6 \times 22$ while the number of channels increases from $1$ to $512$. Subsequently, the generated representations are averaged and a sequence of six upsampling blocks follow with each block containing a transpose convolution layer, leaky ReLU activation and batch normalization. These transpose convolutional blocks increase the spatial dimension of the intermediate representation while decreasing the number of channels from $512$ to $1$, essentially following a hour-glass architecture. The final block has also a sigmoid activation layer that produces the simulated occupancy grid $\hat{V}_{n+1}$ having a $32 \times 32 \times 64$ resolution. Similarly to the classification model Adam optimizer is used. The loss function is define as,

\begin{equation}
L_S = \underbrace{\sum_{i=1}^{B} \|\hat{V}_{n+1}^i - V_{n+1}^i\|_2}_\text{$L_2$ norm} - \alpha \underbrace{\sum_{i=1}^{B}\sum_{j=1}^{M} y_{ij}log(v_{ij})}_\text{cross entropy}
\label{eq:sim-loss}
\end{equation} 
following the same notation as Equation \ref{eq:cla-loss}. It consists of two terms with the first one being the $L_2$ loss between the estimated and actual occupancy grid. The second term is the cross entropy loss between the ground truth classification vector of occupancy grid $V_{n+1}^i$ and the predicted one using the classification model of Figure \ref{fig:3dcnn-clas}. Essentially, the first term pushes the simulated and actual occupancy grids to be spatially similar penalizing any dissimilarity between the ground truth and the predicted occupancy grid. The second term enforces the simulation results to be interpretable by the classification network with the trade-off being controlled by parameter $\alpha$. While the first term ensures that the predicted occupancy grid is visually similar to the ground truth the second term ensures that structural details that determine the class of a grid are preserved. As is shown in the following section the cross entropy term is crucial in generating simulation results that can be successfully used for defect prediction.

\section{Experimental evaluation}
\label{sec::experiments}
This section is devoted to the evaluation of the proposed method. We use the \textit{PCB scans} dataset \cite{PCB_scans} that is also used in \cite{RNet} but instead of estimating glue quantity we predict the 3D structure of the next glue dispension. For the sake of completion we first describe the targeted use case from the microelectronics domain along with the specifications of the used dataset and then proceed with the experimental evaluation of our method.

\subsection{Use-case description}
\label{sec::use-case}

A defect-prone process in microelectronics is the dispensation of conductive glue on an LCP substrate previous to the attachment of a silicon die (IC). The quantity of dispensed glue is the critical parameter that needs to be monitored as insufficient or excessive glue causes defects on the circuit and deteriorate its robustness. Typically, glue quantity is controlled by the pressure on the glue dispenser but is also affected by glue deposits that accumulate over time on the dispenser, the type and viscosity of the glue as well as environmental parameters such as temperature and humidity. Therefore it is crucial to monitor the quantity of dispensed glue and predict any deviations from the nominal values of glue quantity.

\begin{figure}
\centering
\subfloat[]{\includegraphics[width=0.3\linewidth]{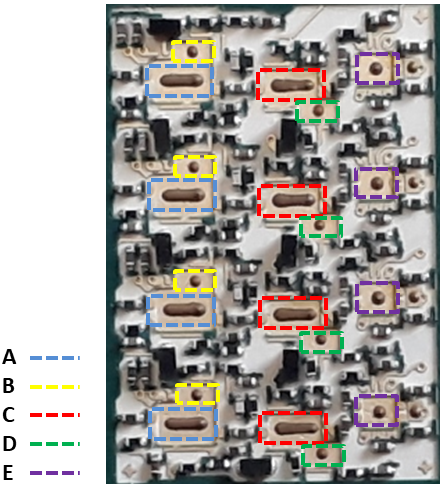}\label{fig:circ-annot}}
\hfil
\subfloat[]{\includegraphics[width=0.5\linewidth]{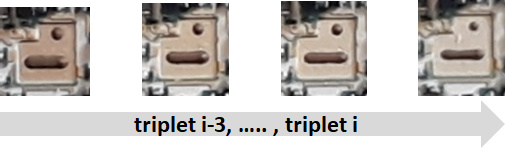}\label{fig:circ-seq}}
\caption{(a) An indicative circuit from the dataset. In has $20$ placeholders that correspond to five different types of ICs, namely $\{A,B,C,D,E\}$. Placeholders of the same type form a column on the circuit. On each placeholder we can see the deposited glue that approximately has the shape of either an ellipse or sphere. From all circuits the third glue deposition is used for the test set and the rest for training. (b) For each triplet of circuits glue deposition are almost identical. In this figure, we see snapshots from $4$ consecutive triplets where the decrease in glue volume is evident. Note that the leftmost glue deposition corresponds to the point cloud of Figure \ref{fig:quan}.}
\end{figure}

In our experiments, we use laser scans from $27$ circuits in the \textit{PCB scans} dataset \cite{PCB_scans} as the one depicted in Figure \ref{fig:circ-annot}. Each circuit is $13mm$ by $19mm$ and has $20$ placeholders where glue is dispensed and five different types $\{A,B,C,D,E\}$ of silicon dies are going to be attached. As is also shown in Figure \ref{fig:circ-annot}, there are four of these placeholders for each type in a circuit. The dispensed glue has different nominal quantity values for each type and has the shape either of an ellipse for types $\{A,C\}$ or a small sphere resembling a dot for $\{C,D,E\}$. 

As is explained in greater detail in \cite{RNet} for each triplet of circuits the glue dispensation process has been manually inspected and controlled so as to have the same quantity of glue deposited for the same type of placeholders. Since each circuit has four placeholders of each type, for each circuit triplet there are $12$ glue deposits with, to the extent possible, identical quantities of glue. Between circuit triplets the pressure on the glue dispenser has been reduced and subsequently the quantity of glue decreases from one circuit triplet to the other. This is also visualized in Figure \ref{fig:circ-seq}, where indicative glue deposits of type $A$ and $B$ are shown from four consecutive circuit triplets. Upon visual inspection we see that the quantity of glue decreases from left to right.

\subsection{Dataset description}
For our experiments we use the point clouds of glue after they are discretized in $32 \times 32 \times 64$ occupancy grids following the process described in \ref{sec::dataparam}. Since there are nine triplets of circuits we respectively define the classes of glue quantity levels with the first class having the most glue and the quantity subsequently decreasing to the next classes. Therefore for each type $\{A,B,C,D,E\}$ there are in total $108$ point clouds divided to nine different classes of glue quantity. Following the same augmentation procedure with \cite{RNet}, we extract $100$ point clouds out of each one having in total $10800$ point clouds for each type. We divide our dataset in a training and test set using a $3$ to $1$ ratio. Specifically as is also explained in Figure \ref{fig:circ-annot} point clouds from the third row of each circuit are assigned to the test set and from rows one, two and four to the training set.

\subsection{Experimental results}

We first evaluate the classification model of Section \ref{sec::nets} that is also graphically summarized in Figure \ref{fig:3dcnn-clas}. We have trained five different instances of the classification model for $20$ epochs using random initialization with a batch size of $64$. Each instance corresponds to a type of IC placeholder. The number of epochs were chosen as no significant improvements was noticed past this point. Classification performance for each of the nine classes and for each type are summarized in Table \ref{tab::class_acc} were precision, recall and f-score are recorded. The last two columns contain the mean performance on classes \textit{V-IX} and on all classes as well. We particularly focus on the last $5$ classes as they are also used for testing, in terms of defect prediction, of simulation results. In all cases, classification performance is well above the prior of each class and verifies that the classification model learns to discriminate the different levels of glue. There is a drop in accuracy for type $D$ and $E$. This is mainly attributed to the small size of glue deposit in these types that cannot be easily captured by the laser scanner. 

\begin{table*}
\tiny
\centering
\begin{tabular}{|c||c|c|c|c|c|c|c|c|c|c|c|c|}
\hline
\textit{Type} & \textit{Metric} & \textit{I} & \textit{II} & \textit{III} & \textit{IV} & \textit{V} & \textit{VI} & \textit{VII} & \textit{VIII} & \textit{IX} & \textit{Mean V-IX} & \textit{Mean} \\
\hline
\hline
\multirow{3}{*}{\textit{A}} & \textit{Precision} & 1.00 & 0.94 & 0.98 & 0.90 & 0.96 & 0.96 & 0.69 & 0.94 & 0.95 & \textit{\textbf{0.90}} & \textit{\textbf{0.92}} \\
\cline{2-13}
& \textit{Recall} & 0.99 & 0.98 & 0.87 & 0.99 & 0.81 & 0.73 & 0.93 & 0.91 & 1.00 & \textit{\textbf{0.88}} & \textit{\textbf{0.91}} \\
\cline{2-13}
 & \textit{F-score} & 0.99 & 0.96 & 0.92 & 0.94 & 0.88 & 0.83 & 0.79 & 0.93 & 0.97 & \textit{\textbf{0.88}} & \textit{\textbf{0.91}} \\
\hline
\multirow{3}{*}{\textit{B}} & \textit{Precision} & 0.96 & 0.93 & 0.94 & 0.89 & 0.68 & 0.69 & 0.89 & 0.96 & 0.89 & \textit{\textbf{0.82}} & \textit{\textbf{0.87}} \\
\cline{2-13}
& \textit{Recall} & 0.96 & 0.95 & 0.95 & 0.86 & 0.91 & 0.51 & 0.85 & 0.86 & 0.96 & \textit{\textbf{0.82}} & \textit{\textbf{0.87}} \\
\cline{2-13}
 & \textit{F-score} & 0.96 & 0.94 & 0.95 & 0.88 & 0.78 & 0.59 & 0.87 & 0.91 & 0.93 & \textit{\textbf{0.81}} & \textit{\textbf{0.87}} \\
\hline
\multirow{3}{*}{\textit{C}} & \textit{Precision} & 0.98 & 0.82 & 0.79 & 0.81 & 0.72 & 0.52 & 0.62 & 0.92 & 0.98 & \textit{\textbf{0.75}} & \textit{\textbf{0.80}} \\
\cline{2-13}
& \textit{Recall} & 0.92 & 0.86 & 0.67 & 0.70 & 0.62 & 0.51 & 0.86 & 0.98 & 1.00 & \textit{\textbf{0.79}} & \textit{\textbf{0.79}} \\
\cline{2-13}
 & \textit{F-score} & 0.95 & 0.84 & 0.73 & 0.75 & 0.67 & 0.52 & 0.72 & 0.95 & 0.99 & \textit{\textbf{0.77}}& \textit{\textbf{0.79}} \\
\hline
\multirow{3}{*}{\textit{D}} & \textit{Precision} & 0.67 & 0.70 & 0.74 & 0.30 & 0.37 & 0.40 & 0.60 & 0.88 & 0.73 & \textit{\textbf{0.60}} & \textit{\textbf{0.60}} \\
\cline{2-13}
& \textit{Recall} & 0.85 & 0.48 & 0.47 & 0.21 & 0.60 & 0.48 & 0.50 & 0.65 & 0.98 & \textit{\textbf{0.64}} & \textit{\textbf{0.58}} \\
\cline{2-13}
 & \textit{F-score} & 0.75 & 0.57 & 0.57 & 0.25 & 0.46 & 0.44 & 0.55 & 0.75 & 0.84 & \textit{\textbf{0.61}} & \textit{\textbf{0.57}} \\
\hline
\multirow{3}{*}{\textit{E}} & \textit{Precision} & 0.82 & 0.60 & 0.51 & 0.73 & 0.78 & 0.60 & 0.38 & 0.75 & 0.97 & \textit{\textbf{0.70}} & \textit{\textbf{0.68}} \\
\cline{2-13}
& \textit{Recall} & 1.00 & 0.55 & 0.59 & 0.61 & 0.58 & 0.84 & 0.35 & 0.64 & 0.94 & \textit{\textbf{0.67}} & \textit{\textbf{0.68}} \\
\cline{2-13}
 & \textit{F-score} & 0.90 & 0.57 & 0.54 & 0.66 & 0.67 & 0.70 & 0.36 & 0.69 & 0.95 & \textit{\textbf{0.68}} & \textit{\textbf{0.67}} \\
\hline
\end{tabular}
\caption{Classification Precision, Recall and F-Score on the test set.}
\label{tab::class_acc}
\end{table*}

To the best of our knowledge, the task of defect prediction in industrial application using 3D convolutional networks has not been examined in the past, thus there are no obvious competitors that we can compare with the proposed simulation method. Therefore, we examine two variants of the simulation model. In the first one training is performed exclusively using $L2$, setting the $\alpha$ parameters of Equation \ref{eq:sim-loss} to zero while in the second we set $\alpha$ to $0.1$. We use as input to the simulation network four occupancy grids corresponding to consecutive circuit triplets and the network produces an estimate of the next occupancy grid. For a training specimen we randomly sample an occupancy grid from each circuit triplet. Concretely, a training sample consists of five occupancy grids from five consecutive triplets with the last one being the ground-truth and the rest the inputs to our network. To generate the training set we use a sliding window over the triplets, meaning that we select training samples from triplets $(1,..,5),(2,..,6),\dots,(5,..,9)$. Following this process we generate $20000$ training samples that are equally distributed to each window, therefore we have $4000$ training samples per circuit triplet window. The test set consists of $5000$ samples that are generated following the same sampling strategy as in the training set. Similarly to the classification case, test samples are exclusively selected from the third row of each circuit. As before, there are different instances of the model for each glue type that are trained for $20$ epochs using a batch size of $16$. To validate the performance of the simulation network we forward its occupancy grid predictions to the classification model and record whether it can correctly discern its class. 

\begin{figure*}
\centering
\includegraphics[width=0.85\linewidth]{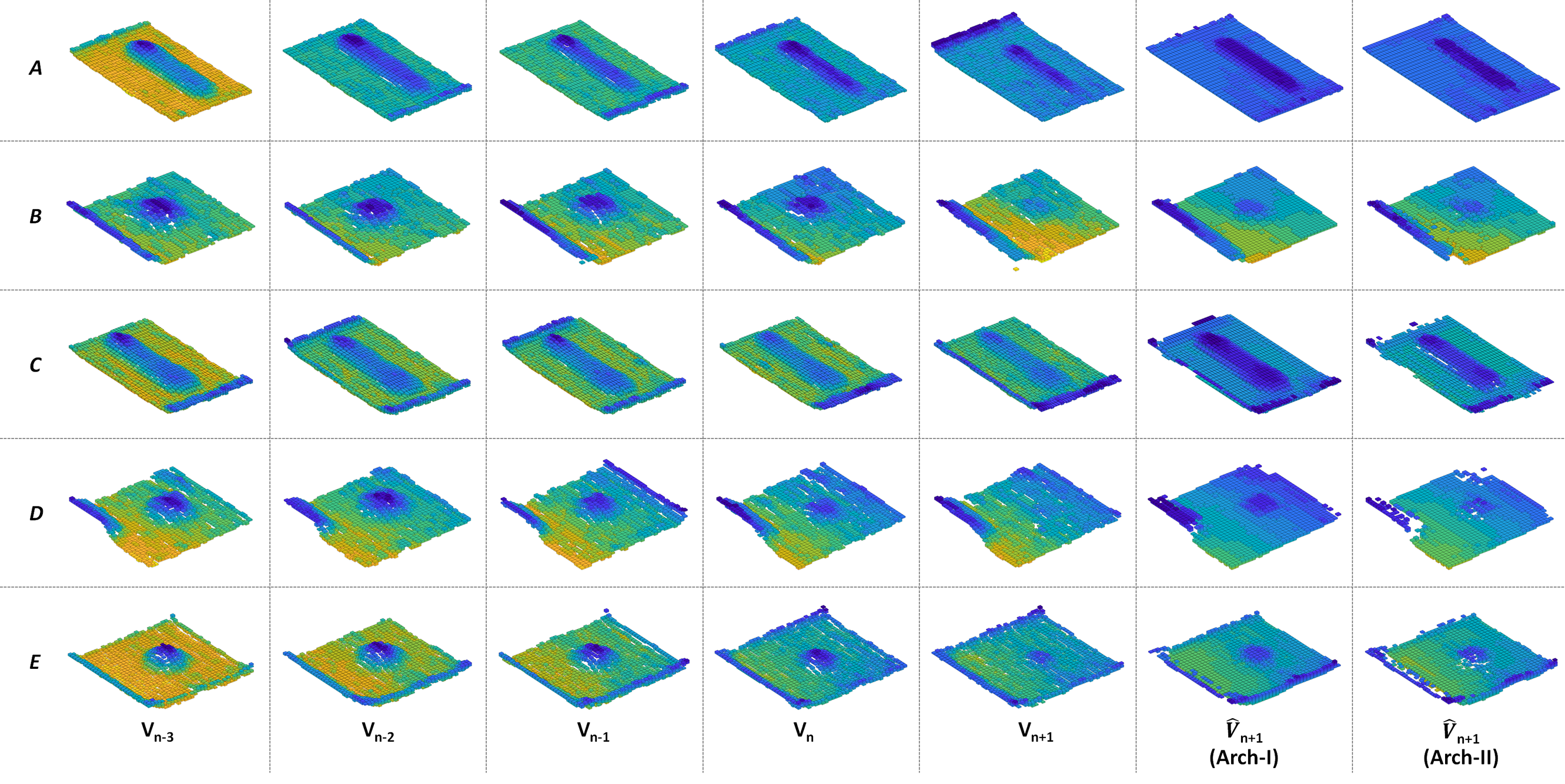}
\caption{Some indicative simulation results from the test set where each row shows an example for a different type of glue placeholder. The first four columns are occupancy grids from consecutive circuit triplets that are provided as input to the simulation network, whereas the fifth is the targeted one from the next triplet as is saved in the test set. The last two columns are the predictions of the simulation network when setting $\alpha = 0.0$ and $\alpha = 0.1$. In the first case we exclude the cross entropy term in Equation \ref{eq:sim-loss}. Notice that in all cased the drop in glue quantity is capture while including the cross entropy term makes the change more sharper as is more evident for types $A$,$C$ and $E$.}
\label{fig:sim-results}
\end{figure*}

Some qualitative results for  different model instances are shown in Figure \ref{fig:sim-results}. In each row, occupancy grids $V_{n-3}, V_{n-2}, V_{n-1}, V_{n}$ are point clouds from samples of four consecutive circuit triplets, while $V_{n+1}$ is a sample from the next triplet. There are in total five rows, one for each type of glue placeholder. As is also illustrated in Figure \ref{fig:circ-seq} there is an obvious decrease in glue quantity for each type. The last two columns show simulation results on $V_{n+1}$, denoted as $\hat{V}_{n+1}$ for the two examined variants where we either exclude the cross entropy loss in Equation \ref{eq:sim-loss} during training  or we include it by setting $\alpha=0.1$. Notice that in both cases the simulated occupancy grid is visually close to the actual, nonetheless in the second case the drop in glue quantity is more evident.

From the perspective of an industrial application, it is more interesting to examine whether the classification model can understand correctly the class of the simulated occupancy grids. This way we can predict any defects related to the change in glue quantity enabling for instance the timely maintenance of the glue dispenser. Results of this evaluation process on the entire test set are reported in Table \ref{tab::pred_class}, where precision, recall and f-score are included. Since our network has as input four past occupancy grids, evaluation is performed only for classes $V$ to $IX$. First of all, classification accuracy is in all cases significantly better than random prediction meaning that the simulation model generates meaningful predictions of future glue deposits. More interestingly the addition of the cross-entropy term  in the loss function for the second network variant dramatically improves classification performance on simulated data, providing effective prediction of future glue levels. It should be also noted that classification performance on simulation results is on par or even exceeds the accuracy  on ground-truth occupancy grids, thus demonstrating the importance of the supervisory signal that the classification network provides during the training of the simulation model.

\begin{table}
\small
\centering
\begin{tabular}{|c|c|c|c|c|c|c|c|c|}
\hline
\textit{Type} & \textit{Metric} & \textit{Net} & \textit{V} & \textit{VI} & \textit{VII} & \textit{VIII} & \textit{IX} & \textit{Mean} \\
\hline
\hline
\multirow{6}{*}{\textit{A}} & \multirow{2}{*}{\textit{Precision}} & \textit{arch-1} & 1.00 & 0.49 & 0.75 & 0.81 & 0.39 & \textit{0.69}\\
\cline{3-9}
& & \textit{arch-2} & 0.98 & 0.86 & 0.98 & 0.81 & 0.79 & \textit{\textbf{0.88}}\\
\cline{2-9}
 & \multirow{2}{*}{\textit{Recall}} & \textit{arch-1} & 0.05 & 0.74 & 0.33 & 0.31 & 1.00 & \textit{0.49}\\
\cline{3-9}
& & \textit{arch-2} & 0.84 & 0.97 & 0.82 & 0.74 & 1.00 & \textit{\textbf{0.87}}\\
\cline{2-9}
 & \multirow{2}{*}{\textit{F-score}} & \textit{arch-1} & 0.09 & 0.59 & 0.46 & 0.45 & 0.56 & \textit{0.43}\\
\cline{3-9}
& & \textit{arch-2} & 0.90 & 0.91 & 0.90 & 0.77 & 0.88 & \textit{\textbf{0.87}}\\
\cline{2-9}
\hline
\multirow{6}{*}{\textit{B}} & \multirow{2}{*}{\textit{Precision}} & \textit{arch-1} & 0.00 & 0.49 & 0.32 & 0.00 & 0.52 & \textit{0.27}\\
\cline{3-9}
& & \textit{arch-2} & 1.00 & 0.60 & 0.95 & 0.77 & 1.00 & \textit{\textbf{0.86}}\\
\cline{2-9}
 & \multirow{2}{*}{\textit{Recall}} & \textit{arch-1} & 0.0 & 0.59 & 1.00 & 0.0 & 0.37 & \textit{0.39}\\
\cline{3-9}
& & \textit{arch-2} & 0.33 & 1.00 & 1.00 & 0.95 & 0.71 & \textit{\textbf{0.80}}\\
\cline{2-9}
 & \multirow{2}{*}{\textit{F-score}} & \textit{arch-1} & 0.00 & 0.53 & 0.49 & 0.00 & 0.43 & \textit{0.29}\\
\cline{3-9}
& & \textit{arch-2} & 0.50 & 0.75 & 0.97 & 0.85 & 0.83 & \textit{\textbf{0.78}}\\
\cline{2-9}
\hline
\multirow{6}{*}{\textit{C}} & \multirow{2}{*}{\textit{Precision}} & \textit{arch-1} & 1.00 & 0.00 & 0.01 & 0.00 & 0.20 & \textit{0.24}\\
\cline{3-9}
& & \textit{arch-2} & 1.00 & 0.71 & 0.73 & 0.87 & 0.91 & \textit{\textbf{0.84}}\\
\cline{2-9}
 & \multirow{2}{*}{\textit{Recall}} & \textit{arch-1} & 0.02 & 0.00 & 0.00 & 0.00 & 1.00 & \textit{0.20}\\
\cline{3-9}
& & \textit{arch-2} & 0.71 & 0.68 & 0.87 & 0.90 & 1.00 & \textit{\textbf{0.83}}\\
\cline{2-9}
 & \multirow{2}{*}{\textit{F-score}} & \textit{arch-1} & 0.05 & 0.00 & 0.00 & 0.00 & 0.34 & \textit{0.08}\\
\cline{3-9}
& & \textit{arch-2} & 0.83 & 0.69 & 0.79 & 0.88 & 0.95 & \textit{\textbf{0.83}}\\
\cline{2-9}
\hline
\multirow{6}{*}{\textit{D}} & \multirow{2}{*}{\textit{Precision}} & \textit{arch-1} & 0.00 & 0.00 & 0.27 & 0.00 & 0.55 & \textit{0.16}\\
\cline{3-9}
& & \textit{arch-2} & 0.92 & 0.54 & 0.79 & 0.61 & 0.70 & \textit{\textbf{0.71}}\\
\cline{2-9}
 & \multirow{2}{*}{\textit{Recall}} & \textit{arch-1} & 0.00 & 0.00 & 0.94 & 0.00 & 0.84 & \textit{0.35}\\
\cline{3-9}
& & \textit{arch-2} & 0.57 & 0.52 & 0.84 & 0.83 & 0.69 & \textit{\textbf{0.69}}\\
\cline{2-9}
 & \multirow{2}{*}{\textit{F-score}} & \textit{arch-1} & 0.00 & 0.00 & 0.42 & 0.00 & 0.66 & \textit{0.22}\\
\cline{3-9}
& & \textit{arch-2} & 0.70 & 0.53 & 0.81 & 0.70 & 0.70 & \textit{\textbf{0.69}}\\
\cline{2-9}
\hline
\multirow{6}{*}{\textit{E}} & \multirow{2}{*}{\textit{Precision}} & \textit{arch-1} & 0.91 & 0.00 & 0.00 & 0.64 & 0.24 & \textit{0.36}\\
\cline{3-9}
& & \textit{arch-2} & 0.71 & 0.81 & 0.95 & 0.82 & 0.70 & \textit{\textbf{0.80}}\\
\cline{2-9}
 & \multirow{2}{*}{\textit{Recall}} & \textit{arch-1} & 0.33 & 0.00 & 0.00 & 0.31 & 1.00 & \textit{0.33}\\
\cline{3-9}
& & \textit{arch-2} & 0.87 & 0.32 & 0.84 & 0.91 & 0.96 & \textit{\textbf{0.78}}\\
\cline{2-9}
 & \multirow{2}{*}{\textit{F-score}} & \textit{arch-1} & 0.49 & 0.00 & 0.00 & 0.42 & 0.39 & \textit{0.26}\\
\cline{3-9}
& & \textit{arch-2} & 0.78 & 0.46 & 0.89 & 0.86 & 0.81 & \textit{\textbf{0.76}}\\
\cline{2-9}
\hline
\end{tabular}
\caption{Precision, Recall, F-Score on simulation. In all cases, the network variant arch-2, where the cross-entropy term is included in the loss function, performs significantly better.}
\label{tab::pred_class}
\end{table}

\section{Conclusions}
\label{sec:conclusions}
The defect prediction system that is presented in this paper allows the simulation of geometrical changes in 3D structures and the detection of defects on simulated results towards the prediction of defective states. The backbone of the system consists of a simulation 3DCNN model that based on previous 3D point clouds estimates the next one as well as a classification 3DCNN model that essentially discriminates between defective and normal states.

The experimental evaluation of the method has been performed on 3D point clouds from the glue dispensation process in microelectronics and has shown promising results for the prediction of defects. A significant finding of our work is that the additional supervisory signal from the classification model during the training of the simulation network is crucial to achieve satisfactory prediction accuracy.

Although experimental results are promising, the proposed system has certain limitations that need to be addressed before shop-floor deployment and application in other industrial use cases. One such limitation is the automated registration of consecutive 3D scans which was straightforward in the examined microelectronics use case but is very challenging for more complex and deformable 3D objects. Moreover, the gradual drop in glue quantity is not characteristic for other industrial use cases with more abrupt changes and remains to be seen how well the proposed methodology would generalize. Finally, another limiting factor is the resolution of the 3D occupancy grids which can be prohibitive for monitoring larger regions with high accuracy.
\section*{Acknowledgment}

This work has been partially supported by the European Commission through project Z-Fact0r funded by the European Union H2020 programme under Grant Agreement no. 723906. The opinions expressed in this paper are those of the authors and do not necessarily reflect the views of the European Commission.
\section*{References}
\bibliographystyle{elsarticle-num}
\bibliography{SMPT2019}

\end{document}